\documentclass[a4paper,twoside]{article}

\usepackage{epsfig}
\usepackage{subcaption}
\usepackage{calc}
\usepackage{amssymb}
\usepackage{amstext}
\usepackage{amsmath}
\usepackage{amsthm}
\usepackage{multicol}
\usepackage{pslatex}
\usepackage{apalike}
\usepackage[bottom]{footmisc}
\usepackage{SCITEPRESS}     % Please add other packages that you may need BEFORE the SCITEPRESS.sty package.

\usepackage{amsmath}
\usepackage{amssymb}
\usepackage{mathtools}
\usepackage{amsthm}

\theoremstyle{definition}
\newtheorem{definition}{Definition}

\begin{document}

%\title{Bridging Causality and Multi-Agent Reinforcement Learning: Causation as a Stimulator for Independent Agents}

\title{Learning Independently from Causality in Multi-Agent Environments}

\author{\authorname{Rafael Pina\sup{1}\orcidAuthor{0000-0003-1304-3539}, Varuna De Silva\sup{1}\orcidAuthor{0000-0001-7535-141X} and Corentin Artaud\sup{1}}
\affiliation{\sup{1}Institute for Digital Technologies, Loughborough University London, 3 Lesney Avenue, London, United Kingdom}
\email{\{r.m.pina, v.d.de-silva, c.artaud2\}@lboro.ac.uk}
}


\abstract{Multi-Agent Reinforcement Learning (MARL) comprises an area of growing interest in the field of machine learning. Despite notable advances, there are still problems that require investigation. The lazy agent pathology is a famous problem in MARL that denotes the event when some of the agents in a MARL team do not contribute to the common goal, letting the teammates do all the work. In this work, we aim to investigate this problem from a causality-based perspective. We intend to create the bridge between the fields of MARL and causality and argue about the usefulness of this link. We study a fully decentralised MARL setup where agents need to learn cooperation strategies and show that there is a causal relation between individual observations and the team reward. The experiments carried show how this relation can be used to improve independent agents in MARL, resulting not only on better performances as a team but also on the rise of more intelligent behaviours on individual agents.}

\onecolumn \maketitle \normalsize \setcounter{footnote}{0} \vfill

\section{\uppercase{Introduction}}
\label{sec:introduction}
The use of causality in the field of Artificial Intelligence (AI) has been gaining the attention of the research community. Recent discussions argue how causality can play an important role to improve many traditional machine learning approaches \cite{peters_2017_elements}. More specifically, recent works argue that causality can be used to get a deeper understanding of the underlying properties of systems within the field of AI. While it can be relatively straight forward to learn the underlying distributions of a given system, to understand the cause of the events in the environments can be a key to a richer representation of the dynamics of the system \cite{peters_2017_elements}. For instance, humans apply causal reasoning in their everyday lives, whereas AI entities are currently incapable of such reasoning. A popular example in the healthcare field is when a certain model trained to make predictions mistakes correlations for causations. If the model is predicting patient health needs based on many factors, it is likely that it will be doing it based on some correlation found. However, in certain cases some of the correlations found might not explain the prediction for all the cases, since the factor found would not be the cause of the prediction, although they were somehow correlated \cite{sgaiersemak_2020_the}.

Motivated by the indications of how causality can be so successfully linked to machine learning, the applications have been studied in different fields. In neurology, causality has been used to find causal relations among different regions of the brain \cite{glymour_2019}. This can be important to understand the reason of certain events lighted by different parts of our brains. Besides the relevance in the healthcare field, in agriculture it can be critical to understand what is causing the harvesting to be less fruitful in one year than in the previous, and not only to see what is correlated to this event \cite{sgaiersemak_2020_the}.

Applications involving more advanced machine learning methods have been rising. For instance, in \cite{kipf_neural_2018} the authors enlighten how causal relations can also be found in human bodies to understand relations among joints when moving. When it comes to time series analysis, the applications are numerous. Starting from the foundations of Granger Causality \cite{granger_1969}, causal discovery in time series has evolved quickly with notable progress. Causal discovery has proved to be helpful in time series prediction related to areas such as weather forecasting or finance \cite{katerina_2007}. More advanced techniques have also shown how the basic concepts of the traditional statistics causality can be integrated with deep learning methods. A popular approach is to use an encoder-decoder architecture to model the causal relations of a certain system \cite{Zhu2020Causal,lowe_amortized_2022,huang_causal_rl}. Alternatives inspired by Granger Causality have also been presented. For instance, the application of Transfer Entropy \cite{schreiber_measuring_2000} is seen as a popular alternative for causal discovery in time series. In fact, this metric is so much related to the traditional Granger Causality that it has been proved that the two metrics are equivalent for Gaussian variables \cite{granger_te_gauss}. 

Despite the advances made with causal relations in machine learning, finding these relations is still challenging \cite{sgaiersemak_2020_the}. Most machine learning models have problems to recognize the patterns necessary to say if two events are correlated or if instead one is the cause of the other. Prompted by the inspiring discoveries in the field, we intend to show how we can use causal estimations within the context of AI in the aims of solving complex cooperative tasks. In particular, we aim to demonstrate how causal estimations can be used in the field of Multi-Agent Reinforcement Learning (MARL) and train these AI agents to work as a team and solve cooperative tasks. MARL is a growing topic in the field of machine learning with many challenges yet to address \cite{canese_multi-agent_2021}. In this sense, the goal of this work is to show how causal estimations can be used to improve independent learners in MARL, by tackling a well-known problem in the field: the lazy agent problem \cite{sunehag_value-decomposition_2018}. By creating the bridge between MARL and causal estimations we hope to open doors to future works in the field, demonstrating how beneficial the link of causal estimations and MARL can be to develop more intelligent and capable entities.

\section{\uppercase{Background}}
\subsection{Decentralised Partially Observable Markov Decision Processes (Dec-POMDPs)}
In this work we consider Dec-POMDPs \cite{oliehoek_a_concise_2016}, defined by the tuple $\langle S,A,O,P,Z,r,\gamma,N\rangle$, where $S$ and $A$ represent the state and joint action spaces, respectively. We consider a setting where each agent $i\in \mathcal{N}\equiv\{1,\dots,N\}$ has an observation $o_i\in O(s,i):S\times \mathcal{N}\rightarrow Z$. Each agent keeps an action-observation history $\tau_i \in T:(Z\times A)^*$, on which it conditions a stochastic policy $\pi_i(a_i|\tau_i)\rightarrow [0,1]$. At each time step, each agent $i$ takes an action $a_i \in A$ forming a joint action $a=\{a_1,\dots,a_N\}$. Taking the joint action at a state $s$, will make the environment transit to a next state $s'$ according to a probability function $P(s'|s,a):S\times A \times S\rightarrow [0,1]$. All the agents in the team share a reward $r(s,a):S\times A\rightarrow \mathbb{R}$. Let $\gamma \in [0,1)$ be a discount factor.

\subsection{Independent Deep Q-Learning}
Independent Q-learning (IQL) was introduced by \cite{tan_multi-agent_1993} as one of the foundations in MARL. This approach follows the principles of single-agent reinforcement learning and applies the basics of Q-learning \cite{watkins_technical_1992} for independent learners that update their Q-functions individually with a learning rate $\alpha$ following the rule
\begin{equation}\label{eq:q_up}
    \begin{aligned}
    Q(s,a)=(1-\alpha)&Q(s,a)\\
    &+\alpha\left[r+\gamma\mathop{\mathrm{max}}_{a'}Q(s',a')\right]
\end{aligned}
\end{equation}
Later on, \cite{mnih_human-level_2015} introduce Deep Q-Networks (DQNs), a method that combines deep neural networks with Q-learning, allowing agents to approximate their Q-functions as deep neural networks instead of simple lookup tables as in simple Q-learning. This approach introduces the use of a replay buffer that stores past experiences of the agents, and a target network that aims to stabilise the learning process. The DQN is updated in order to minimise the loss \cite{mnih_human-level_2015}
\begin{multline}\label{eq:dqn_loss}
    \mathcal{L}(\theta)=\mathbb{E}_{b\sim B}\left[\big(r+\gamma\mathop{\mathrm{max}}_{a'}Q(s',a';\theta^-)
    \right.
    \\
    \left.
    -Q(s,a;\theta)\big)^2\right]
\end{multline}
where $\theta$ and $\theta^-$ are the parameters of the Q-network and a target Q-network, respectively, for a certain experience sample $b$ sampled from a replay buffer $B$. \cite{tampuu_multiagent_2015} put together the concepts of IQL and DQN to create independent learners that use DQNs. In this paper we refer to this method as Independent Deep Q-learning (IDQL). 

\begin{figure*}[!t]
    \centering
    \includegraphics[width=\textwidth]{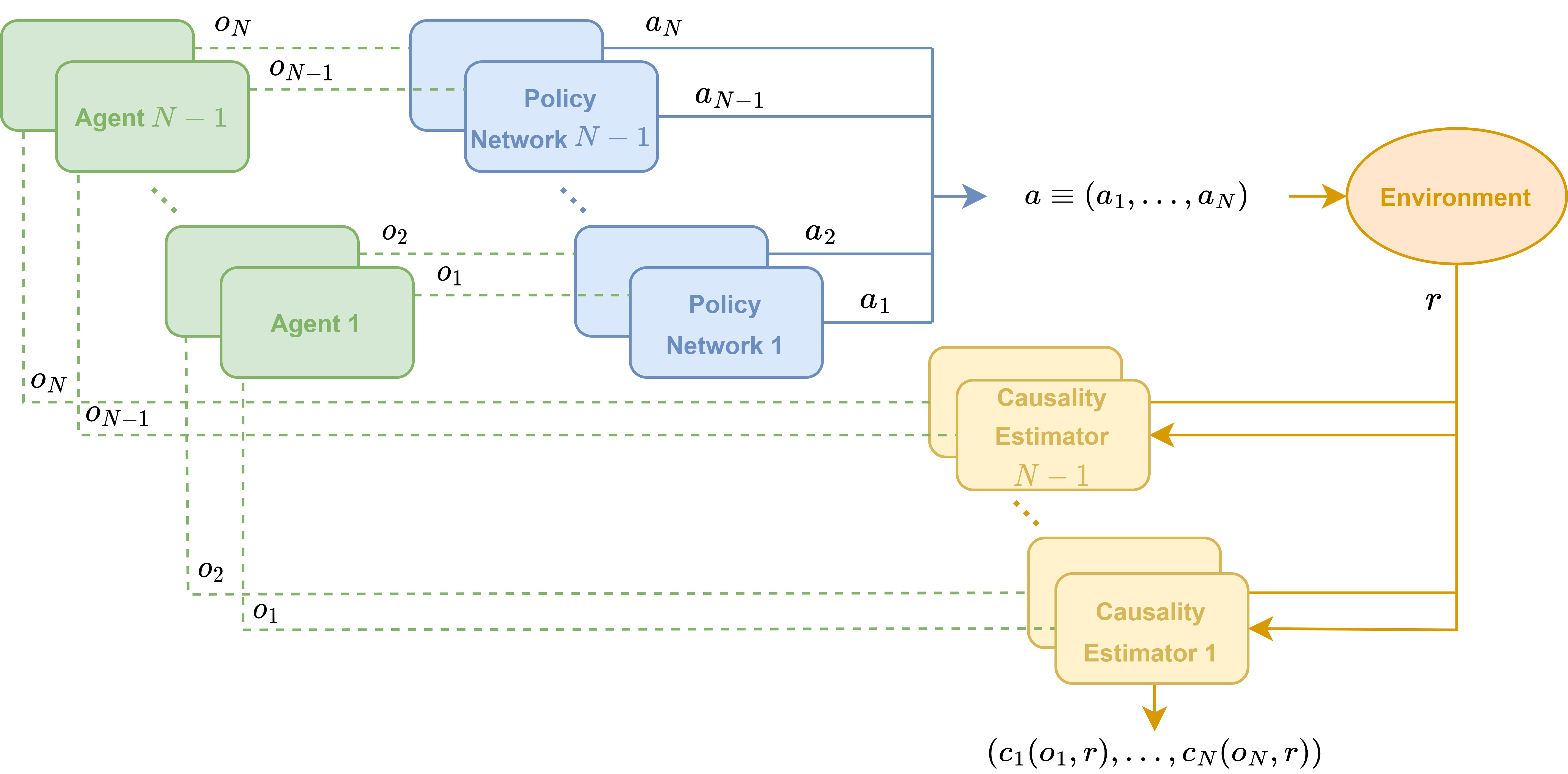}
    \caption{Architecture of the framework described to improve independent learners in MARL using causality detection. Each agent is controlled by an independent network that is updated independently based on the output of the causality estimator, as per Eq. 6.}
    \label{fig:my_label}
\end{figure*}
\subsection{Causality in Time Series}
To build our causality-based approach we used as motivation the concept of Transfer Entropy \cite{schreiber_measuring_2000}. Transfer entropy is a metric that has been widely used to infer about causality relationships in time series for different fields \cite{smirnov_spurious_2013}. Correlation and causality are two concepts that are often misunderstood. Given two time series, while they might be correlated with each other, they may not necessarily be impacted one by another \cite{rohrer_thinking_2018}. Prompted by these concepts, we use the transfer entropy to create the link between causality and MARL scenarios. Definition \ref{def:te} states the formal definition of the Transfer Entropy. 
\begin{definition}\label{def:te}
Given a bivariate time series $y$ with variables $X$ and $Y$, the transfer entropy of $X$ on $Y$ can be expressed as \cite{aziz_transfer_2017}
\begin{equation}
    T_{X\rightarrow Y}=H(Y_t|Y_t^-)-H(Y_t|Y_t^-,X_t^-)
\end{equation}
where $H$ represents the Shannon’s Entropy at a given time stamp $t$ and $X_t^-$ and $Y_t^-$ represent the previous values of the time series, up to $t$.
\end{definition}

\section{\uppercase{Causality in MARL}}
In this section we present the proposed method in this paper, named Independent Causal Learning (ICL). This method aims to punish lazy agents and as a result improve the performance of MARL agents in cooperative tasks. The key for the proposed method is the use of an agent-wise causality factor $c_i$ that each agent $i:i\in\{1,\dots,N\}$ uses to attribute or not the team reward to itself. This way, with the proposed method we intend to improve the credit assignment to the agents in cooperative multi-agent tasks. Each individual agent should be able to understand whether it is helping the team to achieve the intended team goal of the task or not. At the same time, this method encourages the agents to learn only by themselves. This can be beneficial, since in many real scenarios it is often unfeasible to provide the agents with the full information of the environment \cite{canese_multi-agent_2021}. Hence, to learn optimal policies in such scenarios, agents may be forced to rely only on their individual observations to understand whether they are performing well or not. To motivate the proposed method, we explore the concept of temporal causality, commonly used in the context of time series, as stated in Definition \ref{def:te}. Building up from this definition, we can then create the bridge between causality and a MARL problem.
\begin{definition}\label{def:te_2}
Let $E$ represent a certain episode sampled from a replay buffer of experiences. Let $E$ be denoted as a multivariate time series of observations $O$ and rewards $R$. From Definition \ref{def:te}, we can then define the transfer entropy for the time series $E$ as
\begin{equation}
    T_{O\rightarrow R}=H(R_t|R_t^-)-H(R_t|R_t^-,O_t^-)
\end{equation}
where $T$ defines the amount of information reduced in the future values of $R$ by knowing the previous values of $R$ and $O$, and $H$ is the Shannon’s Entropy.
\end{definition}
Definition \ref{def:te_2} states the motivation that supports the existence of a causality relationship between observations and rewards when we see a reinforcement learning episode as a time series. Assuming that these causal relations are indeed present in MARL and can be estimated accurately, we can define a certain function that calculates the causal relations between the team reward and individual observations for each agent $i$,
\begin{equation}\label{eq:causal_eq1}
c_i(o_i,r)=\left\{
\begin{array}{ll}
    1 & o_i\ causes\ r\\
    0 & \lnot\ o_i\ causes\ r
\end{array}, i \in \{1,\ldots,N\}
\right.    
\end{equation}

When this causality factor is calculated accurately, each agent can adjust the team reward received from the environment and update its individual network following the rule
\begin{equation}\label{eq:theo_eq2}
\begin{aligned}
    Q_i(\tau_i,a_i)&=(1-\alpha)Q_i(\tau_i,a_i)\\
    &+\alpha\left[c_i(\tau_i,r)\times r+\gamma\mathop{\mathrm{max}}_{a_i'}Q_i(\tau_i',a_i')\right]
\end{aligned}
\end{equation}

In the considered training setting, the loss calculated by each agent is the same as in IDQL (Eq. 2), but with respect to their individual networks $Q_i$ and following the update in Eq. 6, resulting in the loss,
\begin{multline}\label{eq:icl_loss}
    \mathcal{L}(\theta_i)=\mathbb{E}_{b\sim B}\Big[\big(c_i(\tau_i,r)\times r
    \\
    +\gamma\mathop{\mathrm{max}}_{a_i'}Q_i(\tau_i',a_i';\theta_i^-)-Q_i(\tau_i,a_i;\theta_i)\big)^2\Big]
\end{multline}

\subsection{Causality Effect in MARL Environments}
In this subsection we describe the MARL environment used for the experiments in this paper and show how $c_i$ can be intuitively estimated, when there is some prior knowledge about the task. Fig. \ref{fig:my_label} illustrates the architecture of the proposed method and how the calculation of $c_i$ is incorporated within MARL. Note that the team rewards given result from the sum of $N$ individual rewards.

\begin{figure}
    \centering
    \includegraphics[width=0.8\columnwidth]{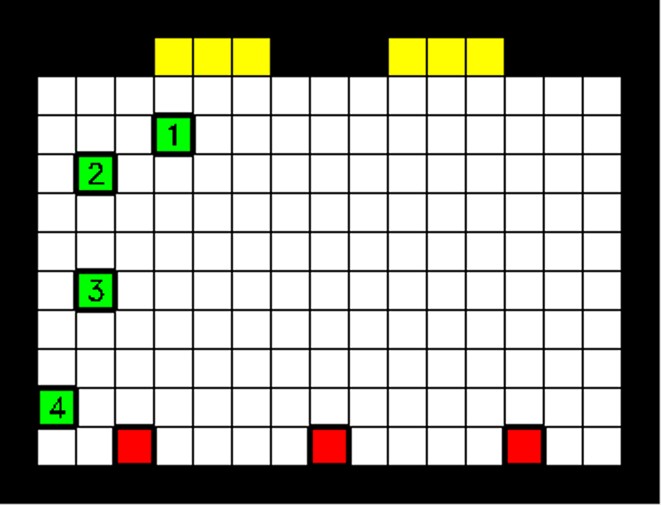}
    \caption{Cooperative Warehouse environment used in the experiments.}
    \label{fig:maps}
\end{figure}

\textbf{Warehouse} Warehouse is a grid world environment with dimensions 10x15 (Fig. \ref{fig:maps}). The goal of this task is to carry boxes from a box delivery queue (red) to a pre-defined dropping station (yellow), simulating a workstation in a real factory of robots with delivery tasks to complete. However, in this environment each box can only be carried by two agents at the same time. In this sense, a team of 4 agents needs to cooperate in order to maximise the number of boxes delivered to the dropping station over the duration of the episodes (300 steps). Every time a box is successfully dropped at the station, each agent receives a reward of +5. Additionally, there is an intermediate reward of +2 for when a box is lifted from the delivery queue. However, in this environment there is also a wrong dropping station that the agents should avoid (left yellow). If the agents drop a box at this fake station, then each one will receive a reward of -5. So, beyond learning to pick the boxes together, the agents should also learn to always choose the right dropping station (right yellow). The observation space for this environment consists of a vector with the position of the agent, a boolean flag $F_i$ that tells whether the agent $i$ is carrying a box or not, and an observation mask with dimensions 5x5 around the agent. In this sense, the causality factors are calculated based on the following condition: there is a positive team reward (box was either picked from queue or delivered), and the agent is carrying a box at the moment right before the reward (box delivered) or the agent was carrying a box at the moment of the reward (box picked),
    \[C_1=
    \left\{
    \begin{array}{ll}
         True& (F_i^{t-1}\vee F_i^{t})\wedge r^t > 0\\
         False& otherwise.
    \end{array}
    \right.
    \]
where $F_i$ is the boolean flag in the observation $o_i$ of an agent $i$ at a timestep $t$. In this scenario, the learning is improved by the perception of each agent about whether they are carrying a box or not at the moment of receiving a team reward. This intuition will coax the agents to be more cooperative with the team, eliminating lazy agents and maximising the number of delivered boxes over the duration of the episodes.

\section{\uppercase{Experiments and Discussions}}
In this section we evaluate the performance of the proposed approach and use IDQL as the benchmark. The motivation behind the use of the baseline IDQL is to demonstrate how the proposed method improves the quality of the behaviours learned for fully independent learners that use only local observations to learn the tasks. Importantly, we investigate the performances of independent learners and how we can improve independent learning. Hence, we do not use the popular parameter sharing convention in MARL \cite{gupta_cooperative_2017} and do not follow the CTDE paradigm \cite{lowe_multi-agent_2017}. In other words, the learning process is fully decentralised and independent.

\subsection{Team Performance and Rewards}
\begin{figure}[!t]
    \centering
    \includegraphics[width=0.7\columnwidth]{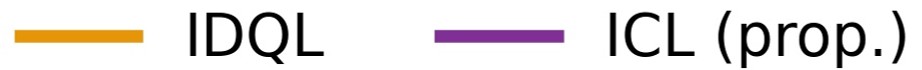}
    \includegraphics[width=\columnwidth]{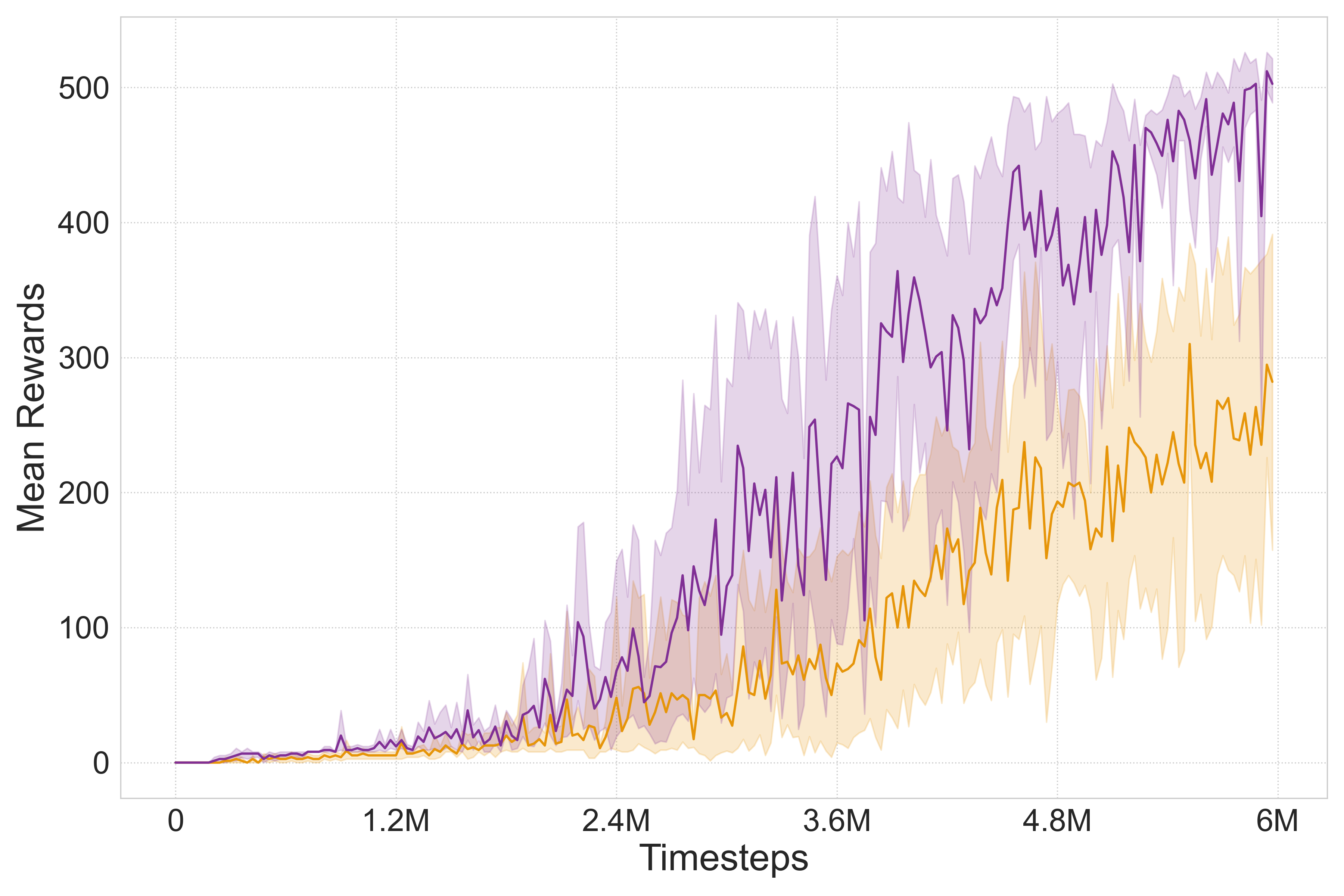}
    \caption{Team rewards obtained for the experimented Warehouse environment (6 independent runs). The bold area represents the 95\% confidence interval.}
    \label{fig:results}
\end{figure}

Fig. \ref{fig:results} illustrates the results for the Warehouse task, where ICL achieves a much higher performance. Although IDQL can still achieve an intermediate reward, it is clear the discrepancy of performance between the two methods. The sub-optimal reward of IDQL is explained by insufficient exploration, caused by lack of cooperation of some agents. As Fig. \ref{fig:results} shows, ICL encouraged the agents to learn different policies and hence to cooperate more, leading to an optimal reward that could not be achieved by the fully independent learners of IDQL. This means that using a causality estimator for independent agents to understand whether they have caused the team rewards or not will benefit the team as a whole, pushing them to develop more cooperative behaviours and not become lazy.

\begin{figure}
    \centering
    \includegraphics[width=\columnwidth]{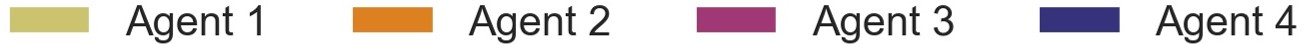}
    \\
    \vspace{1mm} 
    \subfloat[][]{\label{fig:add_res_a}\includegraphics[width=0.8\columnwidth]{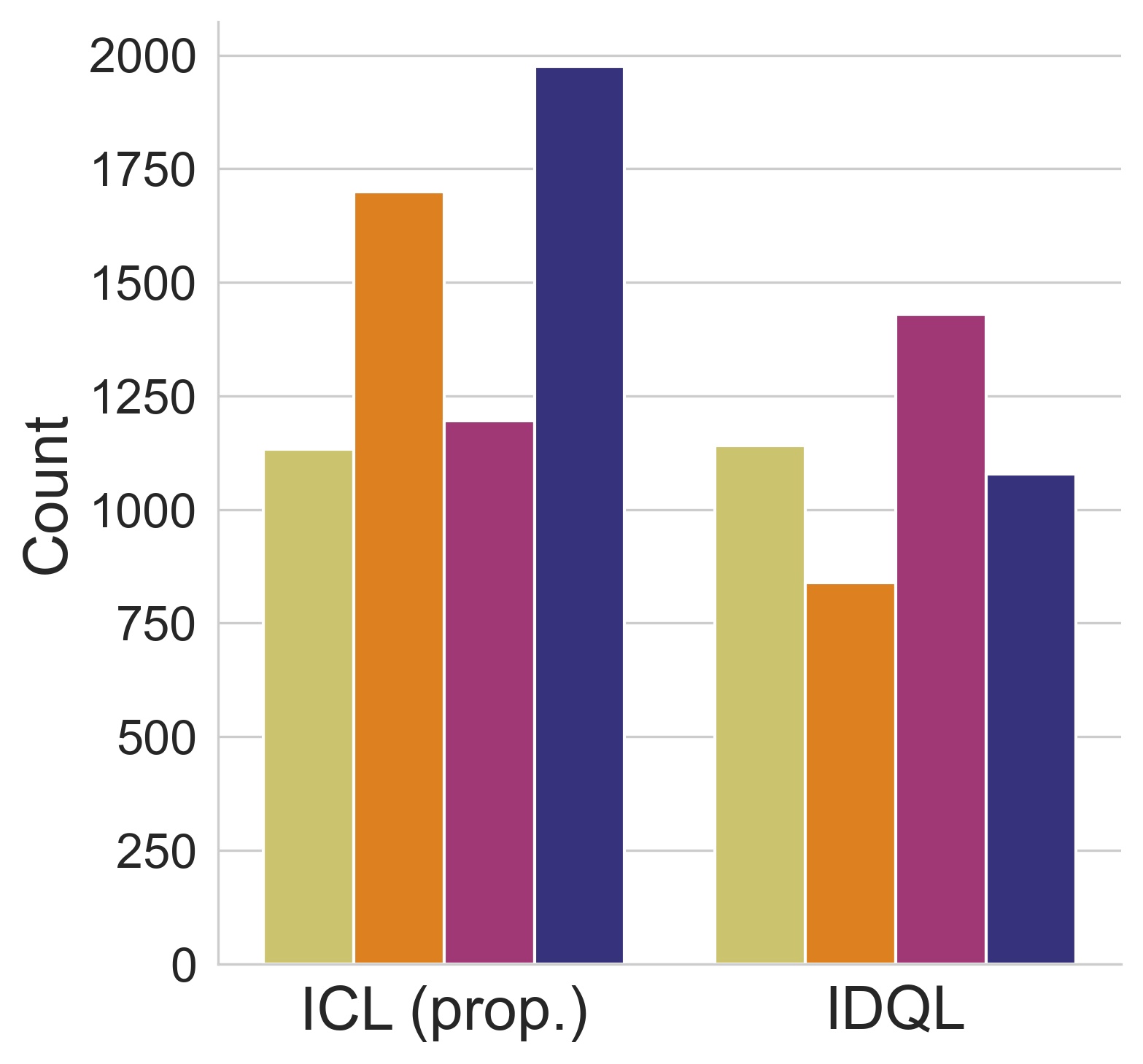}}
    \\
    \vspace{1mm} 
    \subfloat[][]{\label{fig:add_res_b}\includegraphics[width=0.8\columnwidth]{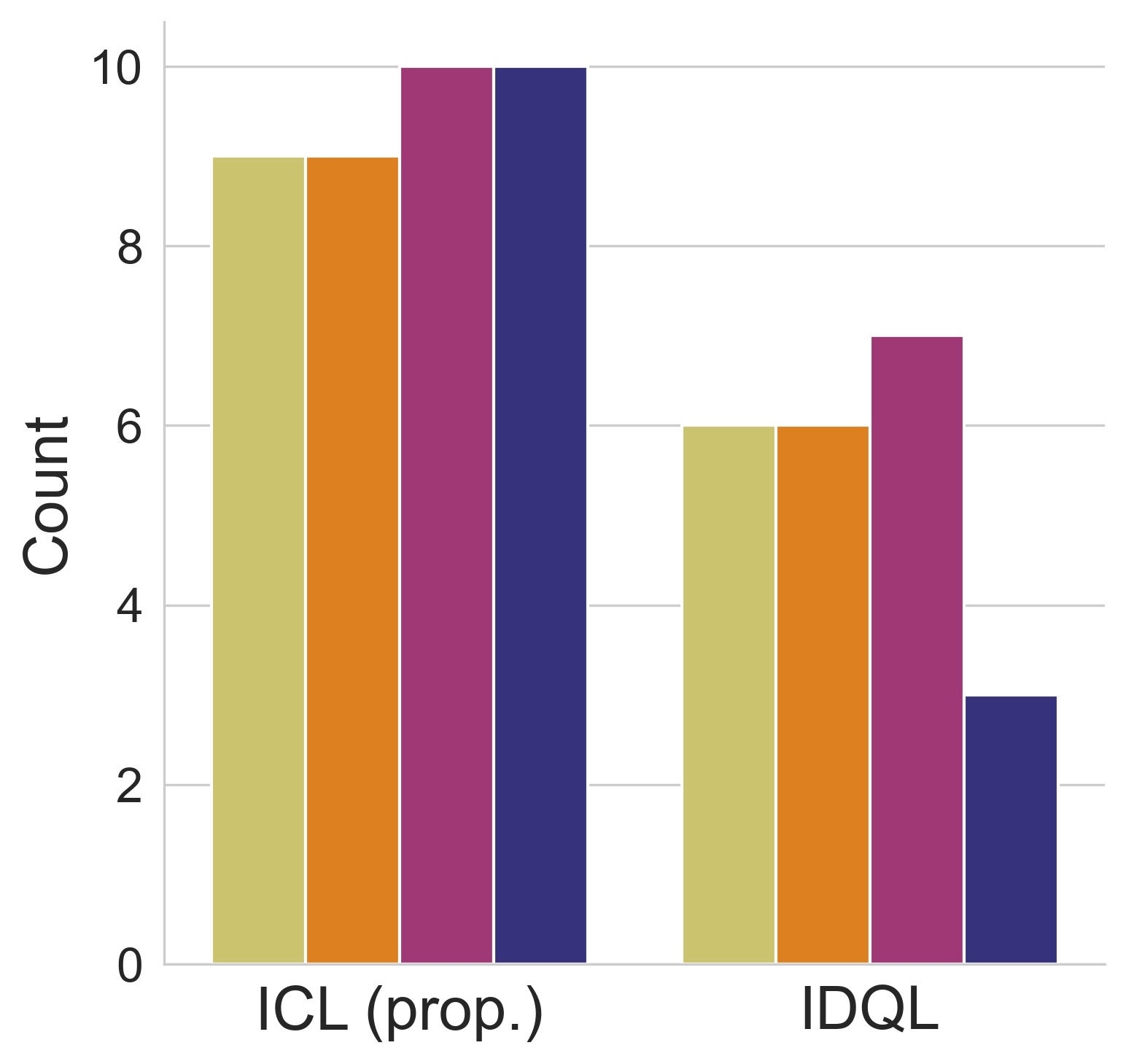}}
    \caption{Behaviour metrics for ICL vs IDQL in the experimented environments. From top to bottom: (a) distance travelled per agent in Warehouse (one successful episode); (b) boxes delivered per agent in Warehouse (one successful episode, same as (a)).}
    \label{fig:add_results}
\end{figure}

\subsection{Individual Learned Behaviours}
We have seen that the agents trained by ICL with a causality estimator can achieve higher rewards as a team in the complex environment experimented. We now investigate the quality of the individual behaviours learned by ICL compared to the purely independent learners of IDQL. The aim of this analysis is to demonstrate how ICL enables independent and individual agents to learn more intelligent behaviours. For the results presented in this subsection, we selected trained policies of IDQL and ICL for the warehouse environment.

As we discuss in detail ahead, Fig. \ref{fig:add_results}a and Fig. \ref{fig:add_results}b show how the proposed method eliminates lazy agents in the warehouse task by looking at how much the agents move during the task and how many boxes they deliver (contribute to the goal of the team). The discrepancy of rewards between IDQL and ICL in Fig. \ref{fig:results} is explained by the existence of agents that do not cooperate and let the others do the task all by themselves. For instance, this can be confirmed in Fig. \ref{fig:add_results}a that shows how some of the agents trained by IDQL do not move nearly as much as the agents trained by ICL. This means that some of them have become lazy and will not be cooperative with the team, waiting for the other agents to solve the task for them. They keep receiving the team reward, but without credit for it. Additionally, Fig. \ref{fig:add_results}b shows that, while for ICL all the agents carry roughly the same amount of boxes, for IDQL some of them almost do not cooperate. As an example, agent 4 barely participates in the task in IDQL, carrying almost no boxes during the episode. As Fig. \ref{fig:results} demonstrates, such non-cooperative individual behaviours will be very harmful for the team as a whole, leading to worst overall outcomes. In addition, agents with such undesirable behaviours would not be qualified to be placed together with a new team of other trained agents. For instance, if we get to train a couple of agents to solve a certain task and want to transfer them to a different team to help other agents, they would not be qualified to do so.

\section{\uppercase{Conclusion and Future Work}}
\label{sec:conclusion}

This paper introduced Independent Causal Learning (ICL), a method for learning fully independent behaviours in cooperative MARL tasks that bridges the concepts of causality and MARL. When there is some prior knowledge of the environment, a causal relationship between the individual observations and the team reward becomes perceptible. This allows the proposed method to improve learning in a fully decentralised and fully independent manner. The results showed that providing an environment dependent causality estimation allows agents to perform efficiently in a fully independent manner and achieve a certain goal as a team. In addition, we showed how using causality in MARL can improve individual behaviours, eliminating lazy agents that are present in normal independent learners and enabling more intelligent behaviours, leading to better overall performances in the tasks. These preliminary results are inspiring as they enlighten the potential of the link between causality and MARL. 

In the future, we aim to study how causality estimations can be used to also improve centralised learning. In addition, although the recognition of patterns that identify causality relations has shown to be challenging in machine learning methods, we intend to extend this method to more cases and show that causality discovery can be generalised to MARL problems. Furthermore, we intend to study how ICL can be applied in real scenarios that require online learning and prohibit excessive trial-and-error episodes due to potentially catastrophic events caused by the learning agents. We believe that this can be a breakthrough for online learning in real scenarios where reliable communication or a centralised oracle is not available, and agents must learn to coordinate independently. At last, we expect that this link can bring more relevant research questions to the field of MARL.

\vfill
\section*{\uppercase{Acknowledgements}}
This work was funded by the Engineering and Physical Sciences Research Council in the United Kingdom (EPSRC), under the grant number EP/T000783/1.

\bibliographystyle{apalike}
%{\small
%\bibliography{example}}

\end{document}